\documentclass{article}
\usepackage{geometry}
\usepackage{url}
\usepackage{amstext}
\usepackage{amssymb}
\usepackage{graphicx}
\usepackage{pgfplots}

\providecommand{\tabularnewline}{\\}
\usepackage{tikz}\usepackage{authblk}

\usetikzlibrary{arrows,shapes,positioning,automata,decorations.pathmorphing,snakes}

\begin{document}

\title{Generalised Discount Functions
\\ applied to a Monte-Carlo AI$\mu$
Implementation}

\date{\vspace{-5ex}}

\author[1]{Sean Lamont\thanks{sean.a.lamont@outlook.com}}
\author[1]{John Aslanides\thanks{john.stewart.aslanides@gmail.com}}
\author[2]{Jan Leike\thanks{leike@google.com}}
\author[1]{Marcus Hutter\thanks{marcus.hutter@anu.edu.au}}
\affil[1]{Research School of Computer Science, Australian National University}
\affil[2]{Google Deepmind, London}
 \affil[2]{Future of Humanity Institute, University of Oxford}

\providecommand{\keywords}[1]{\textbf{\textit{Keywords---}} #1}
\maketitle
\begin{abstract}
In recent years, work has been done to develop the theory of General
Reinforcement Learning (GRL). However, there are few examples demonstrating
the known results regarding generalised discounting.
We have added to the GRL simulation platform AIXIjs the functionality
to assign an agent arbitrary discount functions, and an environment
which can be used to determine the effect of discounting on an agent's
policy. Using this, we investigate how geometric, hyperbolic and power
discounting affect an informed agent in a simple MDP. We experimentally
reproduce a number of theoretical results, and discuss some related
subtleties. It was found that the agent's behaviour followed what
is expected theoretically, assuming appropriate parameters were chosen
for the Monte-Carlo Tree Search (MCTS) planning algorithm.
\end{abstract}


\section{Introduction}

Reinforcement learning (RL) is a branch of artificial intelligence
which is focused on designing and implementing agents that learn how
to achieve a task through rewards. Most RL methods focus on one specialised
area, for example the Alpha-Go program from Google Deepmind which
is targeted towards the board game Go {[}12{]}. General Reinforcement
Learning (GRL) is concerned with the design of agents which are effective
in a wide range of environments. RL agents use a \textit{discount
function} when choosing their future actions, which controls how heavily
they weight future rewards. Several theoretical results have been
proven for arbitrary discount functions relating to GRL agents {[}8{]}.

We present some contributions to the platform AIXIjs\footnote{For a thorough introduction to AIXIjs, \url{aslanides.io/docs/masters_thesis.pdf}} {[}1{]}{[}2{]}, which
enables the simulation of GRL agents for gridworld problems. Being
web-based allows this platform to be used as an educational tool,
as it provides an understandable visual demonstration of theoretical
results. In addition, it allows the testing of GRL agents in several
different types of environments and scenarios, which can be used to
analyze and compare models. This helps to showcase the different strengths
and weaknesses among GRL agents, making it a useful tool for the GRL
community in terms of demonstrating results. Our main work here is
to extend this platform to arbitrary discount functions.
Using this, we then compare the behaviour induced by common discount
functions and compare this to what is theoretically expected.

We first provide the necessary background to understand the experiments
by introducing the RL setup, agent and planning algorithms, general
discounting, and AIXIjs. We then present details of the environment
and agent implementation used for the analyses. Finally, we present
the experiments and the results, along with a discussion for each
function.

\let\aechar\ae 

\section{Background }

\subsection{Reinforcement Learning Setup}

RL research is concerned with the design and implementation of goal-oriented
agents. The characteristic approach of RL is to associate \textit{rewards}
with the desired goal and allow the \textit{agent} to learn the best
strategy for gaining rewards itself through trial and error {[}14{]}.
The agent interacts with an\textit{ environment} by producing an action
$a$, and the environment responds with an observation and reward
pair $(o,r)=e$ which we call a \textit{percept}. The \textit{history}
up to interaction cycle $k$ is given by the string of all actions
and percepts, $a_{1}e_{1}.....a_{k-1}e_{k-1}$. To simplify notation,
this is written as \ae$_{<k}$. Mathematically, an agent's \textit{policy
}is a stochastic function mapping a history to an action, $\pi:(\mathcal{A}\times\mathcal{E})\rightsquigarrow\mathcal{A},$
while an environment is a stochastic map from a history and an action
to a percept, $\mu:(\mathcal{A}\times\mathcal{E})^{*}\times\mathcal{A}\rightsquigarrow\mathcal{E}$,
where $\rightsquigarrow$ is a stochastic mapping. In the context
of adaptive control, Bellman {[}3{]} first introduced equations for
expressing optimal policies in both deterministic and stochastic environments,
including infinite state spaces. Also introduced was the idea of a
\textit{value function}. A value function is how an agent assigns
value to an environment \textit{state} (or a state-action pair), where
value is a measure of the expected future discounted reward sum. To
solve the Bellman equations, it is necessary to assume a fully observable
\textit{Markovian} environment (a \textit{Markov Decision Process},
or a \textit{MDP}). In an MDP, the agent can observe all relevant
information from the environment at any time, without needing to remember
the history. Although useful for MDPs, many problems of interest lack
the necessary assumptions to tractably solve the Bellman equations.
The problem of scaling RL to non-Markovian and \textit{partially observable}
real world domains provides the motivation for General Reinforcement
Learning.

In such cases, it is useful to express the value function in terms
of the agent's history, with the value of a policy $\pi$ with history
$\text{�}_{<t}$ and environment $\mu$ given by the equation:

\begin{equation}
V_{\mu}^{\pi}(\text{�}_{<t}):=\mathbb{\mathbb{E_{\mu}^{\pi}}}\left[\sum_{k=t}^{\infty}\gamma_{k}r_{k}|\text{�}_{<t}\right]
\end{equation}
Where $r$ is the reward and $\gamma$ is a discount function {[}9{]}.
This equation gives the $\mu$-expected utility for a policy $\pi$
. If we are in a MDP, then we can replace the history by the current
state, and rewrite this as a Bellman Equation {[}3{]}.

\subsection{AI$\mu$}

The GRL agent AI$\mu$ {[}4{]} is purposed to find the optimal reward
in a known environment. There are no other assumptions made about
the environment, so this agent extends to partially observable cases.
AI$\mu$ is simply defined as the agent which maximises the value
function given by (1). Specifically, for any environment $\mu$,
\begin{equation}
\pi^{AI\mu}\in\arg\max_{\pi}V_{\mu}^{\pi}
\end{equation}

As there is usually no way to know the true environment, the main
purpose of AI$\mu$ is to provide a theoretical upper bound for the
performance of an agent for a given environment. As we wish to isolate
the effect of discounting, AI$\mu$ is the agent used for our experiments
to remove uncertainty in the agent's model.

\subsection{Generalised Discounting}

A discount function is used to weight rewards based on their temporal
position relative to the current time. There are several motivations
for using a discount function to determine utility, as opposed to
taking an unaltered sum of rewards. In practice, a discount function
allows the agent's designer to decide how it would like the agent
to value rewards based on how far away they are. A discount function
also serves to prevent the utility from diverging to infinity, as
is the case when using undiscounted reward sums.

Samuelson {[}11{]} first introduced the model of discounted utility,
with the utility at time $k$ given by the sum of discounted future
rewards:
\begin{equation}
V_{k}=\sum_{t=k}^{\infty}\gamma_{t-k}r_{t}
\end{equation}
 This model is the most commonly used in both RL and other disciplines,
but has several issues. These include that the discount function cannot
change over time, and that the value of an action is independent of
the history. Hutter and Lattimore {[}8{]} address several issues with
this model first by using the GRL framwork to allow decisions which
consider the agent's history. They also generalise the setting to allow
a change in discounting over time. Specifically, they define a discount
vector $\mathbf{\gamma}^{k}$ for each time step $k$, with the entries
in the vector being the discount applied at each time step $t>k.$
Replacing $\gamma_{t-k}$ with $\mathbf{\gamma}^{k}$ in (3) gives
a more general model of discounted utility, as it allows the discount
function to change over time by using different vectors for different
time steps.

Using this model, Hutter and Lattimore {[}8{]} provide a general classification
of \textit{time inconsistent }discounting. Qualitatively, a policy
is time consistent if it agrees with previous plans and time inconsistent
if it does not. For example, if I plan to complete a task in 2 hours
but then after 1 hour plan to do it after another 2 hours, my policy
will be time inconsistent. Formally, an agent using discount vectors
$\gamma^{k}$ is time consistent iff:
\begin{equation}
\forall k,\;\exists a_{k}>0\;such\;that\enskip\gamma_{t}^{k}=a_{k}\mathbf{\gamma}_{t}^{1},\quad\forall t\geq k\in\mathbb{N}
\end{equation}

Which is to say, the discount applied from the current time $k$ to
the reward at time $t$ is equal to some positive scalar multiple
of the discount used for $t$ at time 1.

Also presented in their work is a list of common discount functions
and a characterisation of which of these are time consistent. These
form the basis for our experiments and we present a taxonomy below:

Given the current time $k$, future time $t>k$, and a discount vector
$\gamma$, we have:

\textit{Geometric Discounting}: $\mathbf{\gamma}_{t}^{k}=g^{t}, g\in(0,1)$.
Geometric discounting is the most commonly used discount function,
as it provides a straightforward and predictable way to value closer
rewards higher. It is also convenient as for $\gamma\in(0,1)$ it ensures the
expected discounted reward (i.e. value) will always be bounded, and
therefore well defined in all instances. Geometric discounting is
always time consistent, which is apparent when considering the definition
in (4).

\textit{Hyperbolic Discounting}: $\mathbf{\gamma}_{t}^{k}=\frac{1}{(1+\kappa(t-k))^{\beta}},\kappa\in\mathbb{R}^{+},\beta\geq1$.
Hyperbolic discounting has been thought to accurately model human
behaviour, with some research suggesting humans discount this way
when deciding actions {[}15{]}. Hyperbolic discounting is time inconsistent,
which is much of the reason why it is considered to model many irrational human
behaviour patterns. It is clear that hyperbolic discounting is time
inconsistent, as it is not possible to factor the above expression
in a way which satisfies (4). Hyperbolic discounting is most commonly
seen for $\beta=1$, with $\beta>1$ ensuring the discounted reward
sum doesn't diverge to infinity.

\textit{Power Discounting}: $\mathbf{\gamma}_{t}^{k}=t^{-\beta},\beta>1$.
Power discounting is of interest because it causes a \textit{growing
effective horizon.} This in effect causes the agent to become more
far sighted over time, with future rewards becoming relatively more
desirable as time progresses. This is flexible as there is no need
to assign an arbitrary fixed effective horizon, it will instead grow
over time. Hutter and Lattimore {[}8{]} point out that this function
is time consistent, which combined with the growing effective horizon
makes it an effective means of agent discounting.

\subsection{Monte-Carlo Tree Search with $\rho$UCT }

Monte-Carlo Tree Search (MCTS) is a planning algorithm designed to
approximate the expectimax search tree generated by (1), which is
usually intractable to fully enumerate. UCT {[}7{]} is a MCTS algorithm
which is effective for Markovian settings. Veness et al. {[}16{]}
extend this to general environments with the $\rho$UCT algorithm.
The algortithm generates a tree comprised of two types of nodes, 'decision'
nodes and 'chance' nodes. A decision node reflects the agents possible
actions, while chance nodes represent the possible environment responses.
A summary of the algorithm is as follows: First, plan forward using
standard Monte-Carlo simulation. Then select an action in the tree
using the UCB action policy; Define a search horizon $m$, maximum
and minumum reward $\beta$ and $\alpha$, value estimate $V'$, and
history $h$, with $T(ha)$ being the number of visits to a chance
node, and $T(h)$ the number of visits to a decision node. Then, for
$T(ha)>0$:

\begin{equation}
a_{UCB}=\arg\max_{a}\frac{1}{m(\beta-\alpha)}V'(ha)+C\sqrt{\frac{\log(T(h))}{T(ha)}}
\end{equation}

If $T(ha)=0$ then the best action will default to $a$. The parameter
$C$ is an exploration constant, which can be modified to control
the likelihood that an agent will take an exploratory action. Veness
et al. {[}16{]} remark that high values of $C$ lead to 'bushy' and
short trees, compared to low values yielding longer and more discerning
trees. Once the best action is selected, the values for each node
are updated backwards to the root to reflect the new action. The primary
strength of this algorithm is that it allows for history based tree
search, by using $\rho$ as the current environment model and planning
based on that.

\subsection{AIXIjs}

We implement our experiments using AIXIjs, a JavaScript platform designed
to demonstrate GRL results. AIXIjs is structured as follows: There
are currently several GRL agents which have been implemented to work
in different (toy) gridworld and MDP environments. Using these, there
are a collection of demos which are each designed to showcase some
theoretical result in GRL and are presented on the web page. Once
a demo is selected, the user can choose to alter some default parameters
and then run the demo. This then begins a batch simulation with the
specified agent and environment for the selected number of time cycles
(a batch simulation runs the whole simulation as one job, without
any interference). The data collected during the simulation is then
used to visualise the interaction. The API allows for anyone to design
their own demos based on current agents and environments, and for
new agents and environments to be added and interfaced into the system.
It also includes the option to run the simulations as experiments,
collecting the data relevant to the simulation and storing it in a
JSON file for analysis.

The source code can be accessed on: \url{https://github.com/aslanides/aixijs}

While the demos can be found at: \url{http://aslanides.io/aixijs/}

or \url{http://www.hutter1.net/aixijs/}

There has been some related work in adapting GRL results to a practical
setting. In particular, the Monte-Carlo AIXI approximation {[}16{]}
successfully implemented a AIXI model using the aforementioned $\rho$UCT
algorithm. This agent was quite successful, even within a \textquotedblleft challenge
domain\textquotedblright{} (a modified Pac-Man game with 1060 possible
states) with the agent learning several key tactics for the game and
consistently improving. This example demonstrated that it is possible
to effectively adapt GRL agents to a practical setting, and is the
basis for the approximation of AI$\mu$ presented here.

Related to the AIXIjs platform is the REINFORCEjs web demo by Karpathy
{[}6{]}. This demo implements Q-Learning {[}17{]} and SARSA {[}10{]}
RL methods in a grid world scenario, as well as deep Q-Learning for
two continuous state settings. The limitation of this example is its
restriction to a small set of environments, with Q-Learning and SARSA
being defined only for Markovian environments. These algorithms do
not extend to more complicated environments or agents, which is addressed
by AIXIjs.

\section{Technical Details}

\subsection{AI$\mu$ Implementation }

The agent used for experiments is an MCTS approximated version of
AI$\mu$. By using AI$\mu$, we are removing any potential uncertainty
in the agent's model which facilitates a more accurate analysis of
the effect of discounting. This agent knows the true environment,
so for a fixed discount this implies that its policy $\pi(s)$ will
stay the same for any particular state, assuming a Markovian environment.

Although we will not be using very large tree depth, enumerating the
expectimax by solving equation (1) is not generally feasible. We instead
use MCTS to approximate the search tree, specifically the $\rho$UCT
algorithm introduced in the background. Although UCT would suffice
in our deterministic setting, $\rho$UCT is the default search algorithm
incorporated into AIXIjs and as such was used without modification.

\subsection{Agent Plan and Time Inconsistency Heuristic}

We determine the agent's plan at time step $k$ by traversing the
tree created by $\rho$UCT, first selecting the highest value decision
node and then choosing the corresponding chance node with the most
number of visits. In the case of the environment used here, each decision
node has only one chance child as it is deterministic. The process
is then repeated up to the maximum horizon reached by the search,
and the sequence of actions taken are recorded as the agent's plan.
The plan is recorded as a numeric string representing the sequence
of actions the agent plans to take. For example, a recorded plan of
000111 indicates the agent plans to first take action '0' three times
in a row and then take '1' three times.

If the action at cycle $k$ is not equal to the action predicted by
the plan at time $k-1$ then we consider this time inconsistent. Formally,
if the following equation is satisfied then the action at $t$ is
time inconsistent:

\[
\pi_{\gamma^{k-1}}(S_{k})\neq\pi_{\gamma^{k}}(S_{k})
\]

Where $\pi_{\gamma^{t}}(S_{t})$ is a policy $\pi$ using discount
vector $\gamma^{t}$ in state $S$ at time $k$ and $\pi_{\gamma^{k-1}}(S_{k})$
is the same policy using an older discount vector $\gamma^{k-1}$.

If this is true, then the action will be time inconsistent. If it
is not true, the action may still be time inconsistent in regards
to older plans. This method is used to prevent false positives, as
the agent plan deep in the search tree is often not representative
due to the cutoff at the horizon.

\subsection{Environment Setup}

The environment we use is a deterministic fully observable
finite state MDP, represented by figure 1. This environment is structured
to provide a simple means to differentiate myopic and far-sighted
agent policies. The idea behind the environment is to give the agent
the option of receiving an instant reward at any point, which it will
take if it is sufficiently myopic. The other option gives a very large
reward only after following a second action for $N$ steps. If the
agent is far-sighted enough, it will ignore the low instant reward
and plan ahead to reach the very large reward in $N$ time steps.
Formally, the agent has 2 actions from state $S_{i}$ : The first
is to go to $S_{0}$ and receive a small instant reward $r_{I}$.
The other takes the agent to $S_{i+1}$ (where $i\in\mathbb{Z}/(N+1)\mathbb{Z}$)
and gives very low reward $r_{0}<\frac{1}{N}r_{I},$ and a large reward
$r_{L}>Nr_{I}$ for $i=N-1$. In figure 1 the straight lines represent
the first action $a_{0}$ while the other lines representing the second
action $a_{1}$.

\section{Experiments}

\subsection{Overview}

In this section we present the experiments for the discount functions,
which were conducted using the AIXIjs experiment API mentioned in
the background. In particular, we will investigate the effect of geometric,
hyperbolic and power discounting on the $\rho$UCT AI$\mu$ model.
The environment used was the instance of figure 1 parametrised by
$N=6,r_{I}=4,r_{0}=0,r_{L}=1000$. We use average reward as our metric
for agent performance. We avoid using total reward as, in this environment,
it is monotonically increasing with respect to time. This would affect
the scale of graphs, which could obscure an agent's behaviour. We
now present the MCTS parameters, after which we detail two specific
policies prior to the experiments which comprise the rest of the section.
We introduce these policies to avoid unnecessary overlap in the analysis
of geometric and hyperbolic discounting, as they displayed very similar
behaviours.

\begin{figure}[h]
\begin{centering}
\begin{tikzpicture}[shorten >=1pt,node distance=1.75cm,on grid,auto] \node[state] (start)   {$\text{S}_0$}; \node[state] (bed0) [right=of start] {$\text{S}_1$}; \node[state] (bed1) [right=of bed0] {$\text{S}_2$}; \node[state] (bed2) [right=of bed1] {$\text{S}_3$}; \node[state] (bedN) [right=of bed2] {$\text{S}_N$}; \path[->]      (start) edge [loop above] node {$r_I$} ()      (bed0) edge [bend left] node {$$} (start)      (bed1) edge [bend left] node {$$} (start)      (bed2) edge [bend left] node {$$} (start)      (bedN) edge [bend left] node {$r_I$} (start); \path [->,decoration={snake,amplitude=.4mm,segment length=1mm,post length=1mm}] (start) edge [decorate] node {$r_0$} (bed0); \path [->,decoration={snake,amplitude=.4mm,segment length=1mm,post length=1mm}] (bed0) edge [decorate] node {$r_0$} (bed1); \path [->,decoration={snake,amplitude=.4mm,segment length=1mm,post length=1mm}] (bed1) edge [decorate] node {$r_0$} (bed2); \path [->,decoration={snake,amplitude=.4mm,segment length=1mm,post length=1mm}] (bed2) edge [decorate] node { $\dots$ $r_L$} (bedN); \path [->,decoration={snake,amplitude=.4mm,segment length=1mm,post length=1mm}] (bedN) edge [decorate,bend right] node {$r_0$} (bed0); \end{tikzpicture}
\par\end{centering}

\centering{}\caption{MDP used to conduct discounting experiments}
\end{figure}
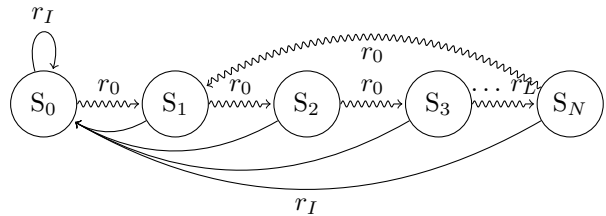

\subsection{MCTS Parameters}

\begin{figure}[h]
\begin{centering}
\begin{tabular}{|c|c|c|c|}
\hline
 & Horizon & UCB Parameter & Samples\tabularnewline
\hline
Geometric & 10 & 0.01 & 10 000\tabularnewline
\hline
\hline
Hyperbolic & 10 & 0.01 & 10 000\tabularnewline
\hline
Power & 7 & 0.001 & 100 000\tabularnewline
\hline
\end{tabular}
\par\end{centering}

\caption{MCTS Parameters used for each discount function}

\end{figure}

It was necessary to increase the samples and lower the exploration
constant for power discounting because over time, the discount factor
becomes exponentially lower with respect to $\beta$. A high exploration
constant would overpower the UCB expression in (5) and result in erratic
policies as there is no clear better action. Given the large number
of samples, it was also necessary to reduce the horizon to shorten
the depth of the tree. 7 is the minimum required to see far enough
into the future to notice the delayed reward.

\subsection{Far-Sighted Policy}

With reference to figure 1, this policy takes action $a_{1}$ (the
alternating arrow) for every time step. The total reward for the far-sighted
policy in 200 time cycles is 33 000, given a delayed reward of 1000
and a reward interval of 6 time steps. Figure 3 presents a plot of
the average reward of this policy in our environment.

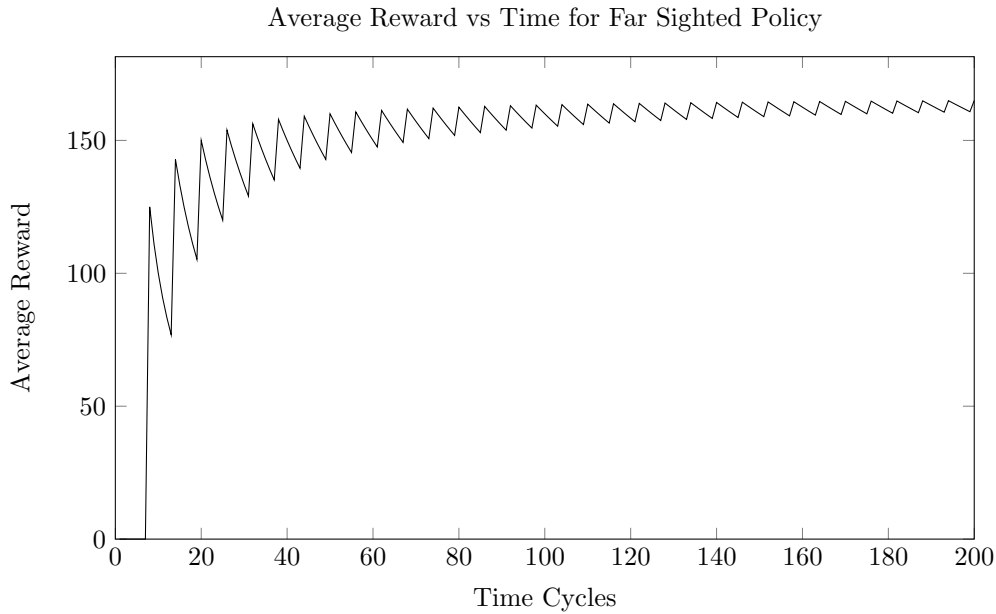
\begin{figure}[h]

\pgfplotsset{width=13cm,height=8cm}

\begin{tikzpicture}

\pgfplotsset{every axis/.append style={ultrathick}
	/tikz/every picture/.append style={
		trim axis left,
		trim axis right,
	}}
	
	\begin{axis}[
	xlabel=Time Cycles,
	ylabel=Average Reward,
	xmin=0,
	xmax=200,
	title=Average Reward vs Time for Far Sighted Policy,
	ymin=0]
	\addplot [mark=none] table{
	1	0
	2	0
	3	0
	4	0
	5	0
	6	0
	7	0
	8	125
	9	111.1111111
	10	100
	11	90.90909091
	12	83.33333333
	13	76.92307692
	14	142.8571429
	15	133.3333333
	16	125
	17	117.6470588
	18	111.1111111
	19	105.2631579
	20	150
	21	142.8571429
	22	136.3636364
	23	130.4347826
	24	125
	25	120
	26	153.8461538
	27	148.1481481
	28	142.8571429
	29	137.9310345
	30	133.3333333
	31	129.0322581
	32	156.25
	33	151.5151515
	34	147.0588235
	35	142.8571429
	36	138.8888889
	37	135.1351351
	38	157.8947368
	39	153.8461538
	40	150
	41	146.3414634
	42	142.8571429
	43	139.5348837
	44	159.0909091
	45	155.5555556
	46	152.173913
	47	148.9361702
	48	145.8333333
	49	142.8571429
	50	160
	51	156.8627451
	52	153.8461538
	53	150.9433962
	54	148.1481481
	55	145.4545455
	56	160.7142857
	57	157.8947368
	58	155.1724138
	59	152.5423729
	60	150
	61	147.5409836
	62	161.2903226
	63	158.7301587
	64	156.25
	65	153.8461538
	66	151.5151515
	67	149.2537313
	68	161.7647059
	69	159.4202899
	70	157.1428571
	71	154.9295775
	72	152.7777778
	73	150.6849315
	74	162.1621622
	75	160
	76	157.8947368
	77	155.8441558
	78	153.8461538
	79	151.8987342
	80	162.5
	81	160.4938272
	82	158.5365854
	83	156.626506
	84	154.7619048
	85	152.9411765
	86	162.7906977
	87	160.9195402
	88	159.0909091
	89	157.3033708
	90	155.5555556
	91	153.8461538
	92	163.0434783
	93	161.2903226
	94	159.5744681
	95	157.8947368
	96	156.25
	97	154.6391753
	98	163.2653061
	99	161.6161616
	100	160
	101	158.4158416
	102	156.8627451
	103	155.3398058
	104	163.4615385
	105	161.9047619
	106	160.3773585
	107	158.8785047
	108	157.4074074
	109	155.9633028
	110	163.6363636
	111	162.1621622
	112	160.7142857
	113	159.2920354
	114	157.8947368
	115	156.5217391
	116	163.7931034
	117	162.3931624
	118	161.0169492
	119	159.6638655
	120	158.3333333
	121	157.0247934
	122	163.9344262
	123	162.601626
	124	161.2903226
	125	160
	126	158.7301587
	127	157.480315
	128	164.0625
	129	162.7906977
	130	161.5384615
	131	160.3053435
	132	159.0909091
	133	157.8947368
	134	164.1791045
	135	162.962963
	136	161.7647059
	137	160.5839416
	138	159.4202899
	139	158.2733813
	140	164.2857143
	141	163.1205674
	142	161.971831
	143	160.8391608
	144	159.7222222
	145	158.6206897
	146	164.3835616
	147	163.2653061
	148	162.1621622
	149	161.0738255
	150	160
	151	158.9403974
	152	164.4736842
	153	163.3986928
	154	162.3376623
	155	161.2903226
	156	160.2564103
	157	159.2356688
	158	164.556962
	159	163.5220126
	160	162.5
	161	161.4906832
	162	160.4938272
	163	159.5092025
	164	164.6341463
	165	163.6363636
	166	162.6506024
	167	161.6766467
	168	160.7142857
	169	159.7633136
	170	164.7058824
	171	163.7426901
	172	162.7906977
	173	161.849711
	174	160.9195402
	175	160
	176	164.7727273
	177	163.8418079
	178	162.9213483
	179	162.0111732
	180	161.1111111
	181	160.2209945
	182	164.8351648
	183	163.9344262
	184	163.0434783
	185	162.1621622
	186	161.2903226
	187	160.4278075
	188	164.893617
	189	164.021164
	190	163.1578947
	191	162.3036649
	192	161.4583333
	193	160.6217617
	194	164.9484536
	195	164.1025641
	196	163.2653061
	197	162.4365482
	198	161.6161616
	199	160.8040201
	200	165
	
	};
	\end{axis}
	\end{tikzpicture}

\caption{Average reward versus time cycles for a far-sighted agent policy}
\end{figure}

The periodic nature of the delayed reward is reflected in the zig
zag shape of the average reward graph. As this policy is consistently
taking $a_{1}$, the time between spikes is constant.

\subsection{Short-Sighted (Myopic) Agent}

The second policy takes action $a_{1}$ (solid arrow in figure 1)
for every time step. The total reward for this policy is 792, given
an instant reward of 4. Figure 4 presents a plot of the average reward
of this policy in our environment.

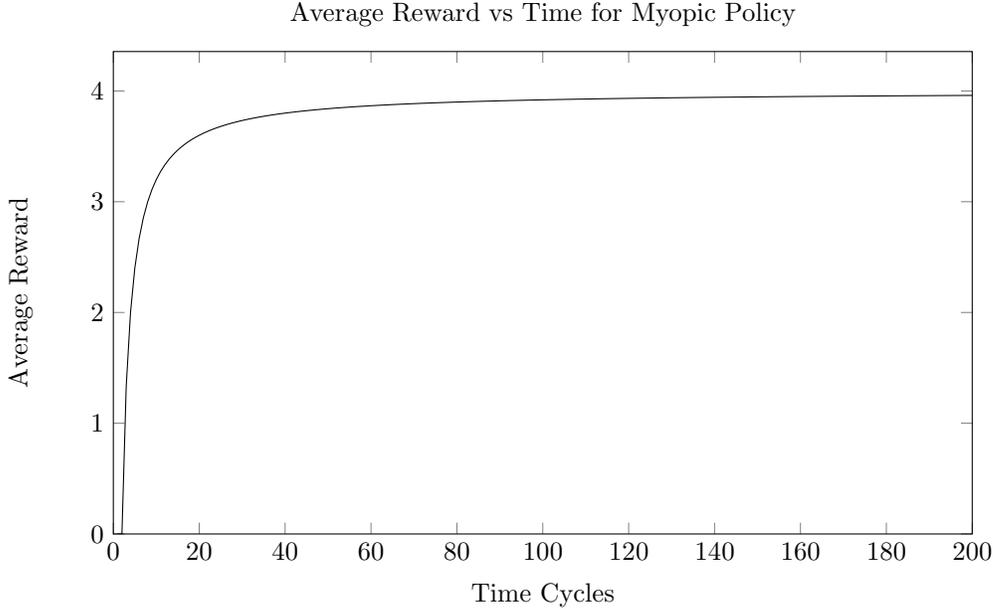
\begin{figure}

\pgfplotsset{width=13cm,height=8cm}

\begin{tikzpicture}

\pgfplotsset{every axis/.append style={ultrathick}
	/tikz/every picture/.append style={
		trim axis left,
		trim axis right,
	}}

\begin{axis}[
     xlabel=Time Cycles,
	 ylabel=Average Reward,
	 xmin=0,
	 xmax=200,
	 title=Average Reward vs Time for Myopic Policy,
	 ymin=0]
\addplot [mark=none] table{
	1	0
	2	0
	3	1.333333333
	4	2
	5	2.4
	6	2.666666667
	7	2.857142857
	8	3
	9	3.111111111
	10	3.2
	11	3.272727273
	12	3.333333333
	13	3.384615385
	14	3.428571429
	15	3.466666667
	16	3.5
	17	3.529411765
	18	3.555555556
	19	3.578947368
	20	3.6
	21	3.619047619
	22	3.636363636
	23	3.652173913
	24	3.666666667
	25	3.68
	26	3.692307692
	27	3.703703704
	28	3.714285714
	29	3.724137931
	30	3.733333333
	31	3.741935484
	32	3.75
	33	3.757575758
	34	3.764705882
	35	3.771428571
	36	3.777777778
	37	3.783783784
	38	3.789473684
	39	3.794871795
	40	3.8
	41	3.804878049
	42	3.80952381
	43	3.813953488
	44	3.818181818
	45	3.822222222
	46	3.826086957
	47	3.829787234
	48	3.833333333
	49	3.836734694
	50	3.84
	51	3.843137255
	52	3.846153846
	53	3.849056604
	54	3.851851852
	55	3.854545455
	56	3.857142857
	57	3.859649123
	58	3.862068966
	59	3.86440678
	60	3.866666667
	61	3.868852459
	62	3.870967742
	63	3.873015873
	64	3.875
	65	3.876923077
	66	3.878787879
	67	3.880597015
	68	3.882352941
	69	3.884057971
	70	3.885714286
	71	3.887323944
	72	3.888888889
	73	3.890410959
	74	3.891891892
	75	3.893333333
	76	3.894736842
	77	3.896103896
	78	3.897435897
	79	3.898734177
	80	3.9
	81	3.901234568
	82	3.902439024
	83	3.903614458
	84	3.904761905
	85	3.905882353
	86	3.906976744
	87	3.908045977
	88	3.909090909
	89	3.91011236
	90	3.911111111
	91	3.912087912
	92	3.913043478
	93	3.913978495
	94	3.914893617
	95	3.915789474
	96	3.916666667
	97	3.917525773
	98	3.918367347
	99	3.919191919
	100	3.92
	101	3.920792079
	102	3.921568627
	103	3.922330097
	104	3.923076923
	105	3.923809524
	106	3.924528302
	107	3.925233645
	108	3.925925926
	109	3.926605505
	110	3.927272727
	111	3.927927928
	112	3.928571429
	113	3.92920354
	114	3.929824561
	115	3.930434783
	116	3.931034483
	117	3.931623932
	118	3.93220339
	119	3.932773109
	120	3.933333333
	121	3.933884298
	122	3.93442623
	123	3.93495935
	124	3.935483871
	125	3.936
	126	3.936507937
	127	3.937007874
	128	3.9375
	129	3.937984496
	130	3.938461538
	131	3.938931298
	132	3.939393939
	133	3.939849624
	134	3.940298507
	135	3.940740741
	136	3.941176471
	137	3.941605839
	138	3.942028986
	139	3.942446043
	140	3.942857143
	141	3.943262411
	142	3.943661972
	143	3.944055944
	144	3.944444444
	145	3.944827586
	146	3.945205479
	147	3.945578231
	148	3.945945946
	149	3.946308725
	150	3.946666667
	151	3.947019868
	152	3.947368421
	153	3.947712418
	154	3.948051948
	155	3.948387097
	156	3.948717949
	157	3.949044586
	158	3.949367089
	159	3.949685535
	160	3.95
	161	3.950310559
	162	3.950617284
	163	3.950920245
	164	3.951219512
	165	3.951515152
	166	3.951807229
	167	3.952095808
	168	3.952380952
	169	3.952662722
	170	3.952941176
	171	3.953216374
	172	3.953488372
	173	3.953757225
	174	3.954022989
	175	3.954285714
	176	3.954545455
	177	3.95480226
	178	3.95505618
	179	3.955307263
	180	3.955555556
	181	3.955801105
	182	3.956043956
	183	3.956284153
	184	3.956521739
	185	3.956756757
	186	3.956989247
	187	3.957219251
	188	3.957446809
	189	3.957671958
	190	3.957894737
	191	3.958115183
	192	3.958333333
	193	3.958549223
	194	3.958762887
	195	3.958974359
	196	3.959183673
	197	3.959390863
	198	3.95959596
	199	3.959798995
	200	3.96
	};
\end{axis}
\end{tikzpicture}

\caption{Average reward versus time cycles for a myopic agent policy}
\end{figure}

The graph reflects the initial reward of 0 as the agent starts off,
and then the constant reward of 4 every following time cycle.

\subsection{Geometric Discounting}

We ran experiments by altering $\gamma$ in increments of 0.1, ranging
from 0.1 to 1.0. We found that in all test runs, the number of time
inconsistent actions given by our heuristic was 0. We found that for $\gamma\leq0.4$ the agent followed exactly the myopic policy from the previous
subsection, receiving a total reward of 792. For $\gamma\geq0.6$
the agent behaved as described in the far-sighted policy subsection,
achieving the optimal reward of 33 000. For $\gamma=0.5$, the agent behaved somewhat erratically,
occasionally altering its behaviour between both policies.
As this value lies between the $\gamma$ which cause strict far/short sightedness,
there would be a small difference in weighted rewards between both policies. It is therefore likely
the erratic behaviour is caused by the MCTS struggling to find the best decision, given there is a degree of
innacuracy in the tree search.  The agent plan enumeration
gave a consistent plan of 0000000000 for $\gamma\leq0.4$ and varied between 1111100000 and 1111111111
for $\gamma\geq0.6$, where '0' and '1' are shorthand for $a_{0}$
and $a_{1}$ respectively. This variation is due to the horizon cutoff at 10, since at some points the agent won't
see far ahead enough to plan for 2 far-sighted actions.

\subsection{Hyperbolic Discounting}

We varied $\kappa$ between 1.0 and 3.0 in increments of 0.2, and kept $\beta$ constant at 1.0. We found
that only $\kappa=1.8$ yielded non-zero time inconsistent actions,
with the total number of such actions recorded as 200. We found for
$\kappa\leq1.8$ that the agent followed exactly the behaviour from
the myopic policy subsection, receiving a total reward of 792. For
$\kappa>1.8$ the agent behaved as described in the far-sighted policy
subsection, achieving the optimal reward of 33 000. The plan for $\kappa=1.8$
was $0111111000$ at every time step. For $\kappa>1.8$ the plan stayed
as 1111111111, and $\kappa\leq1.8$ gave a constant plan of 0000000000.

In the interest of reproducibility, the experiments for hyperbolic
discounting were performed on commit 3911d73 on the provided github
link. The results can also be replicated with recent and future versions,
however the MCTS parameters may need to be adjusted.

\subsection{Power Discounting}

We only used a single $\beta$ value in this case, with $\beta$=
1.01. We note that any change in $\beta$ would result in similar
behaviour, with only the length and time between these stages changing
(hence we need only present the results of one $\beta$ value). No
time inconsistent actions were detected for this function. The total
reward obtained by the agent was 15412.

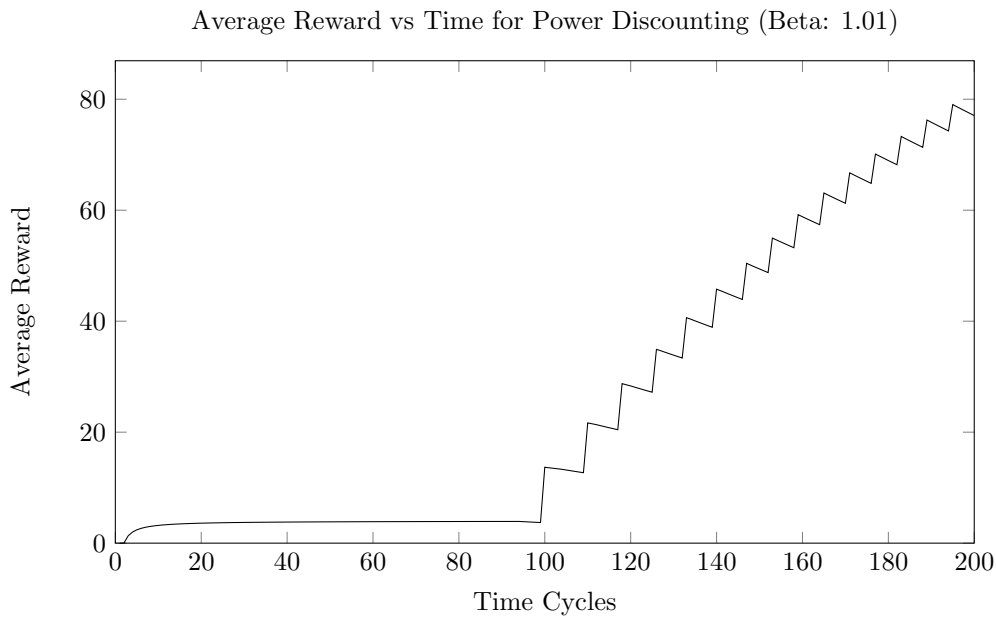
\begin{figure}

\pgfplotsset{width=13cm,height=8cm}

\begin{tikzpicture}

\pgfplotsset{every axis/.append style={ultrathick}
	/tikz/every picture/.append style={
		trim axis left,
		trim axis right,
	}}
	
	\begin{axis}[
	xlabel=Time Cycles,
	ylabel=Average Reward,
	xmin=0,
	xmax=200,
	title=Average Reward vs Time for Power Discounting (Beta: 1.01),
	ymin=0]
	\addplot [mark=none] table{
		1	0
		2	0
		3	1.333333333
		4	2
		5	2.4
		6	2.666666667
		7	2.857142857
		8	3
		9	3.111111111
		10	3.2
		11	3.272727273
		12	3.333333333
		13	3.384615385
		14	3.428571429
		15	3.466666667
		16	3.5
		17	3.529411765
		18	3.555555556
		19	3.578947368
		20	3.6
		21	3.619047619
		22	3.636363636
		23	3.652173913
		24	3.666666667
		25	3.68
		26	3.692307692
		27	3.703703704
		28	3.714285714
		29	3.724137931
		30	3.733333333
		31	3.741935484
		32	3.75
		33	3.757575758
		34	3.764705882
		35	3.771428571
		36	3.777777778
		37	3.783783784
		38	3.789473684
		39	3.794871795
		40	3.8
		41	3.804878049
		42	3.80952381
		43	3.813953488
		44	3.818181818
		45	3.822222222
		46	3.826086957
		47	3.829787234
		48	3.833333333
		49	3.836734694
		50	3.84
		51	3.843137255
		52	3.846153846
		53	3.849056604
		54	3.851851852
		55	3.854545455
		56	3.857142857
		57	3.859649123
		58	3.862068966
		59	3.86440678
		60	3.866666667
		61	3.868852459
		62	3.870967742
		63	3.873015873
		64	3.875
		65	3.876923077
		66	3.878787879
		67	3.880597015
		68	3.882352941
		69	3.884057971
		70	3.885714286
		71	3.887323944
		72	3.888888889
		73	3.890410959
		74	3.891891892
		75	3.893333333
		76	3.894736842
		77	3.896103896
		78	3.897435897
		79	3.898734177
		80	3.9
		81	3.901234568
		82	3.902439024
		83	3.903614458
		84	3.904761905
		85	3.905882353
		86	3.906976744
		87	3.908045977
		88	3.909090909
		89	3.91011236
		90	3.911111111
		91	3.912087912
		92	3.913043478
		93	3.913978495
		94	3.914893617
		95	3.873684211
		96	3.833333333
		97	3.793814433
		98	3.755102041
		99	3.717171717
		100	13.68
		101	13.58415842
		102	13.49019608
		103	13.39805825
		104	13.30769231
		105	13.18095238
		106	13.05660377
		107	12.93457944
		108	12.81481481
		109	12.69724771
		110	21.67272727
		111	21.51351351
		112	21.35714286
		113	21.16814159
		114	20.98245614
		115	20.8
		116	20.62068966
		117	20.44444444
		118	28.74576271
		119	28.53781513
		120	28.33333333
		121	28.09917355
		122	27.86885246
		123	27.64227642
		124	27.41935484
		125	27.2
		126	34.92063492
		127	34.67716535
		128	34.40625
		129	34.13953488
		130	33.87692308
		131	33.61832061
		132	33.36363636
		133	40.63157895
		134	40.35820896
		135	40.05925926
		136	39.76470588
		137	39.47445255
		138	39.1884058
		139	38.90647482
		140	45.77142857
		141	45.4751773
		142	45.15492958
		143	44.83916084
		144	44.52777778
		145	44.22068966
		146	43.91780822
		147	50.42176871
		148	50.08108108
		149	49.74496644
		150	49.41333333
		151	49.08609272
		152	48.76315789
		153	54.98039216
		154	54.62337662
		155	54.27096774
		156	53.92307692
		157	53.57961783
		158	53.24050633
		159	59.19496855
		160	58.825
		161	58.45962733
		162	58.09876543
		163	57.74233129
		164	57.3902439
		165	63.1030303
		166	62.72289157
		167	62.34730539
		168	61.97619048
		169	61.60946746
		170	61.24705882
		171	66.73684211
		172	66.34883721
		173	65.96531792
		174	65.5862069
		175	65.21142857
		176	64.84090909
		177	70.12429379
		178	69.73033708
		179	69.34078212
		180	68.95555556
		181	68.57458564
		182	68.1978022
		183	73.28961749
		184	72.89130435
		185	72.4972973
		186	72.10752688
		187	71.72192513
		188	71.34042553
		189	76.25396825
		190	75.85263158
		191	75.45549738
		192	75.0625
		193	74.67357513
		194	74.28865979
		195	79.03589744
		196	78.63265306
		197	78.23350254
		198	77.83838384
		199	77.44723618
		200	77.06
		
	};
	\end{axis}
	\end{tikzpicture}

\caption{Average reward versus time cycles for power discounting}
\end{figure}

The behaviour in this circumstance follows three stages: For around
100 time steps, the agent behaves completely myopically, reflected
by the small continuous rise in the first half of the graph. The discount
function then reaches a stage where distant rewards are weighted high
enough so that the agent decides to act far-sightedly. For several
time steps, the agent collects the delayed reward then goes to the
instant one for a few cycles. The number of cycles it stays there
gradually decreases until it strictly follows the far-sighted policy.
This can be seen in the graph, as the intervals between peaks are
larger from cycles 100-150 than 150-200 when the agent acts completely
far-sighted.

\section{Discussion}

In regards to time inconsistent agent behaviour, the results were
consistent with theoretical predictions. Geometric discounting was,
for all instances of $\gamma$, time consistent as expected. Somewhat
suprisingly, hyperbolic discounting was time consistent for all measured
$\kappa$ except 1.8 when it was continuously acting inconsistently.
The results of power discounting also lacked any time inconsistent
actions which is expected.

The hyperbolic agent plan of 0111111000 for $\kappa=1.8$ reflects
some interesting behaviour. We can see the agent is planning to stay
at the instant reward for the next time step, and then move off to
collect a delayed reward. But as this plan is the same for all time
steps, the agent continuously stays on the instant reward planning
to do the better long term action later. In effect, the agent is eternally
procrastinating. The fact that this behaviour can be induced with
this function also supports the claim that hyperbolic discounting
can model certain irrational human behaviour. We note the trailing $0$s
are an artifact of the horizon being too low to see far ahead enough
to notice another delayed reward, given the horizon was only 10.

The results of power discounting clearly demonstrate how a growing
effective horizon can effect an agent's policy. Initially the agent
is too short sighted to collect the delayed reward, but over time
this reward becomes more heavily weighted compared to the instant
reward. After some time the agent starts to collect the delayed reward
and soon is fixed to a far-sighted policy. This shows that a growing
effective horizon can cause an agent to collect distant rewards only
after some time has passed, which again reflects what is theoretically
predicted.

There will continue to be new results proven for GRL, so an avenue
for future work would be to demonstrate those results in a similar
fashion to the work presented here. Our contributions to the AIXIjs
framework would allow for this to be done easily for results pertaining
to agent discounting. Other future work would be the development of
practical GRL systems which extend beyond a toy environment, and which
can be used for non-trivial tasks.

\section{Summary}

We have adapted the platform AIXIjs to include arbitrary
discount functions. Using this, we were able to isolate time inconsistent
behaviour and illustrate the effect of the discount function on an
agent's farsightedness. We were able to show it is possible to use
power discounting in a concrete setting to observe the impact of a growing effective
horizon, which influenced the time at which an agent chose to collect
distant rewards. We also demonstrated that hyperbolic discounting
can induce procrastinating behaviour in an agent. Our current framework
now permits a larger class of experiments and demos with general discounting,
which will be useful for future research on the topic.

\end{document}